\documentclass[10pt,twocolumn,letterpaper]{article}

\usepackage{cvpr}
\usepackage{times}
\usepackage{epsfig}
\usepackage{graphicx}
\usepackage{amsmath}
\usepackage{amssymb}
\usepackage{varwidth}
\usepackage{float}
\usepackage{cite}
\usepackage{amsfonts}

\usepackage[breaklinks=true, pdfborder={0 0 1}, bookmarks=false]{hyperref}

\cvprfinalcopy 

\def\cvprPaperID{****} 

\begin{document}

\title{A unified neural network for object detection,  multiple object tracking and vehicle re-identification}

\makeatletter
\def\@fnsymbol#1{\ensuremath{\ifcase#1\or \dagger\or \ddagger\or
   \mathsection\or \mathparagraph\or \|\or **\or \dagger\dagger
   \or \ddagger\ddagger \else\@ctrerr\fi}}
\makeatother

\author{Yuhao Xu\\
VeilyTech, Wuhan, China\\
{\tt\small yuhao.xu@veilytech.com}
\and
Jiakui Wang\thanks{Corresponding author}\\
VeilyTech, Wuhan, China\\
{\tt\small jiakui.wang@veilytech.com}
}

\maketitle

\begin{abstract}

  Deep SORT\cite{wojke2017simple} is a tracking-by-detetion approach to multiple object tracking 
  with a detector and a RE-ID model.
  Both separately training and inference with the two model is time-comsuming.
  In this paper, we unify the detector and
  RE-ID model into an end-to-end network, by adding an additional track branch for
  tracking in Faster RCNN architecture. With a unified network, we are able
  to train the whole model end-to-end with multi loss, which has shown much benefit
  in other recent works.
  The RE-ID model in Deep SORT needs to use deep CNNs to extract feature map from detected
  object images, However, track branch in our proposed network straight make use of
  the RoI feature vector in Faster RCNN baseline, which reduced the amount of calculation.
  Since the single image lacks the same object which is necessary
  when we use the triplet loss to optimizer the track branch, we
  concatenate the neighbouring frames in a video to construct our training dataset.
  We have trained and evaluated our model on AIC19 vehicle tracking dataset, 
  experiment shows that our model with resnet101 backbone can achieve 57.79 \% mAP
  and track vehicle well.

\end{abstract}

\section{Introduction}
Object tracking is a fundamental yet challenging task in computer vision, 
which requires the algorithm to track the object of multi-frames.
Tracking-by-detection has become the leading paradigm in multiple object
tracking due to recent progress in object detection.

Simple online and realtime tracking (SORT)\cite{bewley2016simple} is a tracking-by-detection
framework which first use a powerful CNN detector to detect 
and then use Kalman filtering and Hungarian method to track.
Deep SORT adds a Deep Association Metric and Appearance information to
enhance the track performance.

However, Deep SORT needs to train the detector and RE-ID model separately,
which can't benefit from the multi loss training that has been proven 
significant to improve the performance in many recent works.
Therefore, we jointly train the detector and RE-ID module in a unified 
network by add an additional track branch to standard Faster RCNN 
architecture. As the RE-ID module needs a triplet, which is lacked in a 
single image, to calculate triplet loss, we concatenate the neighbouring
frames of a video thus we can construct a triplet for training the track
branch. Through this intergrated network, we achieve tracking function
and only add a little computation and memory consume based on Faster RCNN.

\section{Related Work}
\textbf{Object Tracking} Object trackers can be divided into TBD (Tracking
by Detection) and DFT (Detection-Free Tracking) and can also be 
divided to Online and Offline by whether use future frames.

SORT is a TBD and online tracker. This method uses a powerful detector 
to detect objects and use Kalman filter and Hungarian algorithm to
track. SORT can acquire high MOTA while keeping over 100 fps.
DeepSORT adds a Deep Association Metric to enhance SORT.
Deep Association Metric use a large Person Re-identification networks
to judge whether two person in two input images are same.

\textbf{Vehicle ReID} Vehicle Re-identification (ReID) has made great 
progress and
achieve high performance in recent years. Vehicle ReID is to used
to distinguish whether two input images contains same vehicle.
Vehicle ReID can be divided into representation learning based method
and metric learning based method.

Representation learning based method model ReID task as a classification
task, network takes two images as input and has two subnets.
The first subnet outputs predicted id of every image and calculate 
classification loss,
The second subnet integrates the feature of two images and outputs
whether two vehicles is same.
Representation learning based method is robust but generalize poorly.

Metric learning based method is to calculate the similarity of two images.
There are many metrics in this methods such as Contrastive loss\cite{varior2016gated}, Triplet
loss\cite{schroff2015facenet}, Quadruplet loss\cite{chen2017beyond} and Margin sample mining loss\cite{xiao2017margin}.
\cite{marin2018unsupervised} use a detector to identifies all vehicles
and then use triplet network to re-identify vehicles.
\cite{kumar2019vehicle} used a batch all triplet loss with batch weighted
to give more importance to the informative harder samples than trivial
samples.

\textbf{Object Detection} Object detectors can be classified as two-stage 
detector and one-stage detector.
The first two-stage detector with deep convolutional
network is R-CNN\cite{girshick2014rich}, which using sliding window on origin image to get 
predefined anchor and resize anchors to fixed size, then use deep
CNNs to classify whether this anchor includes any object, with an extra
regression offsets to refine the locations of predicted bounding box.
SPP-net\cite{he2015spatial} slides window on CNN feature map instead of origin image to avoid
repeated computation and proposes Spatial Pyramid Pooling (SPP) to get 
fixed length vector from feature patch, which improve the performance of 
detection and speed up the inference.
Fast R-CNN\cite{girshick2015fast} uses RoI pooling to replace SPP and use fully connected layer
to classify and regress instead of SVMs, which can be trained end-to-end
with a multi-task loss.
Faster RCNN\cite{ren2015faster} proposed region proposal networks (RPNs) to generate anchors 
with deep convolutional neural networks (CNNs) and then sample positive
anchors and negative anchors with fixed ratio which can decrease the 
inbalance situation between anchors.

YOLO\cite{redmon2016you} is the first one-stage detector, it divides the CNN feature map into 
a grid of fixed size and predicts fixed number of predicted bounding
boxes with its score on every grid. Although YOLO has a relatively poor
detection performance but it speeds up in a large margin reach to realtime.
SSD\cite{liu2016ssd} first multi CNN layers to generate a pyramid of feature maps on final
CNN feature map
and assign anchors of different scale and aspect ratio in different 
pyramid level,
then classify and regress these anchors.
RetinaNet\cite{lin2017focal} use Feature Pyramid Network (FPN) to replace the multi CNN layers
in SSD which intergrate higher and lower feature.
DSSD\cite{fu2017dssd} uses an integration structure similar to FPN in SSD to generate 
integrated feature map.

\section{Our Approach}

\begin{figure}[htp]
  \centering
  \includegraphics[scale=0.8]{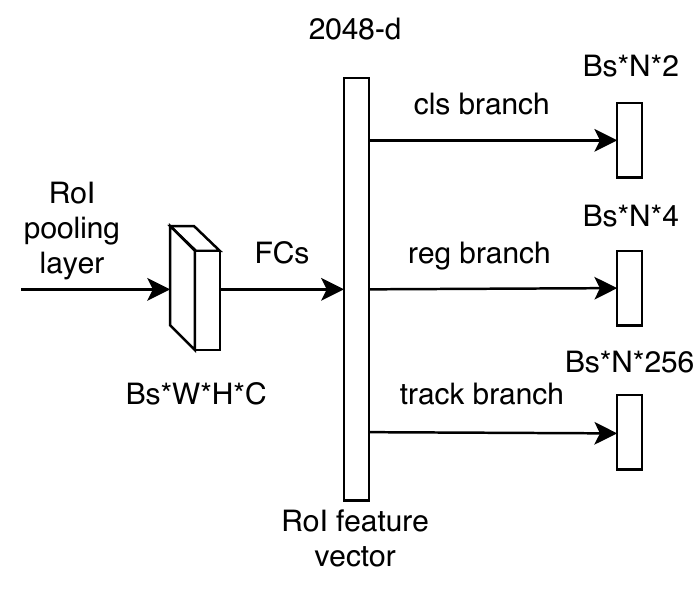}
  \caption{network architecture}
  \label{network}
\end{figure}

\begin{figure*}[htp]
  \centering
  \includegraphics[scale=0.1]{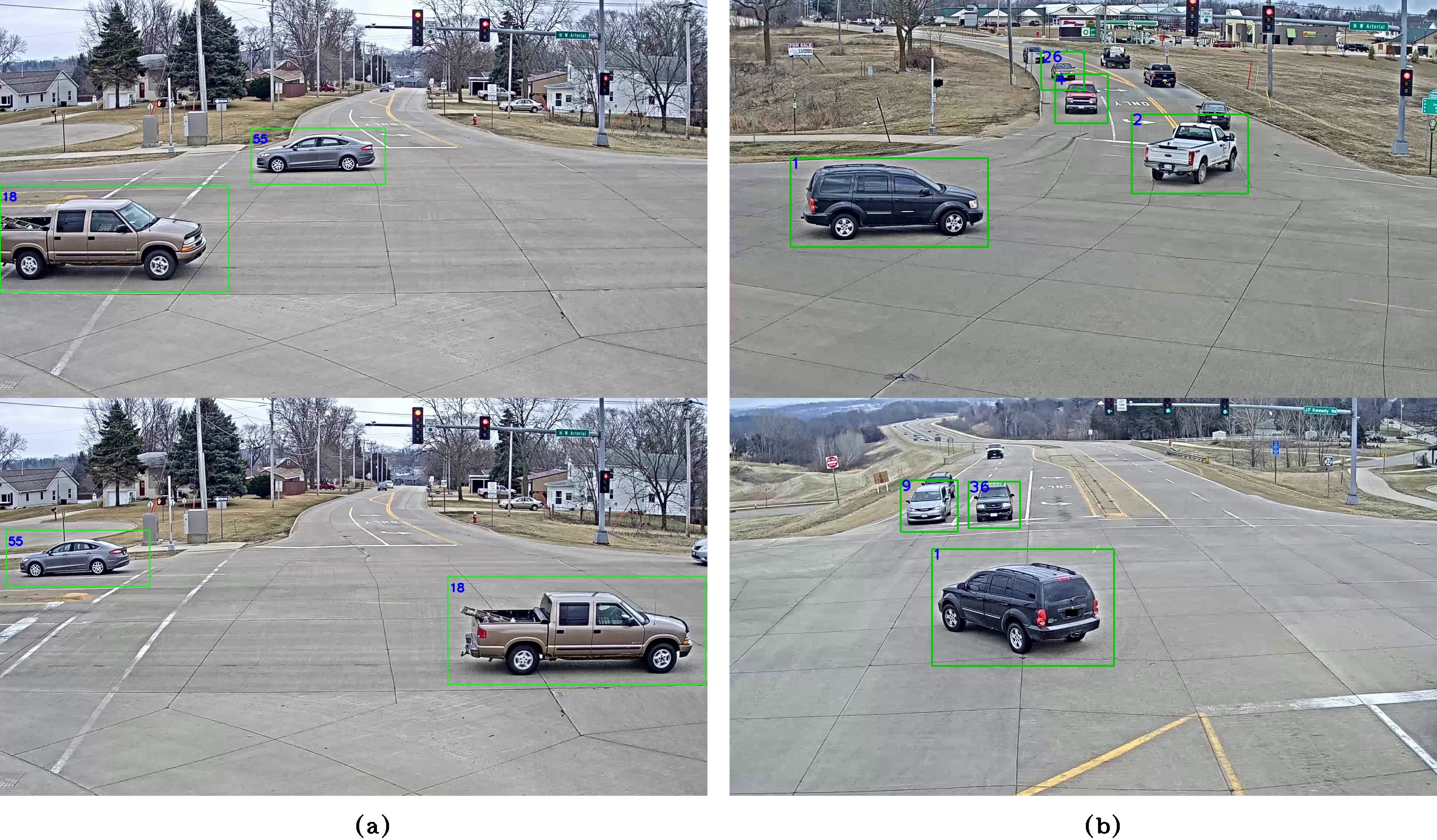}
  \caption{concatenated image for training. \protect \\ (a) is used to train single camera scenario, (b) is used to train multi-camera scenario}
  \label{fig:train}
\end{figure*}

\subsection{Baseline}

This paper uses Faster RCNN as the baseline. To detect objects, 
Faster RCNN first use the Region Proposal Network (RPN) to generate a set of
candidate bounding boxes, each with an objectness score. To generate these
boxes, RPN applies some predifined anchors on the output of last conv layer of
backbone and transforms corresponding feature map to a fixed-length vector, then feeds these vectors into two sibling fully connected layers, one for classification and the other for regression, to output offset and confidence 
score.
Then we select the top anchors and apply
RoI pooling layer and two sibling fully connected layers on corresponding 
feature map to get a RoI feature vector, which followed by
classification branch and regression branch to get the more accurate
classification score and coordinate. Finally, we apply NMS on these 
predict bounding boxes to get final predictions.

\subsection{Track Branch}

Based on Faster RCNN architecture, we propose an additional branch for 
tracking. The net architecture of track branch is the same as classification 
branch, consisting of two fully connection layers but don't share weights.

Fig \ref{network} illustrates the overall network architecture of our 
proposed method. We add a sibling track branch to Faster RCNN.

We use the track branch to extract track feature from the RoI feature vector,
Then use this feature to calculate the distance between different vehicles.
The distance between the same vehicle should be near, and distance between
different vehicle should be relatively far.

We apply the widely used triplet loss to optimize track branch.
\begin{equation}
  \centering
  L_{tri} = max(max(d_{same}) - min(d_{diff}) + m|, 0)
\end{equation}

For every detected bbox, we will calculate the distances between this 
box and other boxes which belong to same vehicles and choose a maximum
distance $max(d_{same})$.
Similarly, minimum distances $min(d_{diff})$ between this box and other 
boxes which belongs to different vehicles will also be calculated.
And then the calculated $\max(d_{same})$ and $\min(d_{diff})$ will be 
used to calculate final triplet loss.
The hyperparam $m$ is used to constrain the distance between 
same vehicle and different vehicle, we set m to 5.0 in this paper.

However, it's hard to use the calculated absolute distance to 
distinguish whether 
the two vehicles belong to the same vehicles or not 
as the triplet loss only ensures the relative distance
between different vehicles. So we use another
``pull"
loss to constrain the distance between same vehicles to be small. It can 
be formulated as:

\begin{equation}
  \centering
  L_{pull} = | max(d_{same}) - m_{pull}|
\end{equation}
where $max(d_{same})$ is the maximum distance between boxes
which belong to the same vehicles.
$m_{pull}$ is the margin to constrain the max same distance, we set $m_{pull}$ to 1.0 in this paper.

When we train the model, we can't find the same vehicles pair in one image,
so we concatenate the two neighbouring frames in a video to construct a
large image to ensure there are same vehicles in one image.
When the input image includes same vehicles, it can be used to jointly
train detection and track tasks. When the input image doesn't include same
vehicles, it can only be used to train the detection tasks since the triplet
loss $L_{tri}$ and ``pull" loss $L_{pull}$ is zero.

Total loss of jointly training is formulated as:

\begin{equation}
  \centering
  Loss = \lambda_{1}L_{cls} + \lambda_{2}L_{reg} + \lambda_{3}L_{tri} + \lambda_{4}L_{pull}
\end{equation}
where $\lambda_{1} \sim \lambda_{4}$ are weights for different loss, 
where $\lambda_{1} \sim \lambda_{2}$ are set to 1.0,
$\lambda_{3} \sim \lambda_{4}$ are set to 0.2.

\subsection{Training}
Before jointly training, 
We first train original Faster RCNN for some epochs and ensure it can detect
objects within acceptable limits. This is because the roi feature vector 
will have large change as detected bounding box get accurate, which may
harm the tracking performance.

While jointly training, detection tranch is trained as usual.
When we train the track branch, we first get the output of detection branch, 
and filter them by threshold $p$, we set $p$ to 0.5.
The IoU of every predicted bounding box and ground truth bounding box will
be calculated and an object id and image id will be assigned to predicted bounding
box if the IoU is larger than 0.5.
If the number of predicted box that which has IoU greater than 0.5 with any ground truth box is more than 1,
, we only choose the box which has highest
IoU and abandon other boxes to ensure every object id in one image is assigned to only one predicted
bounding box. The image id is 0 if the corresponding
ground truth bbox belong to the first image in concatenated image,1
otherwise. Predicted bounding boxes with IoU less than 0.5 with all ground
truth bbox is abandoned.
After assigning every predicted bbox, we can now calculate the $L_{tri}$ and
$L_{pull}$.

We train our network on single GTX 1080ti GPU and use a batch size of 2.
We first train the standard Faster RCNN for 10 epoch and train the jointly 
network for 40 epoch.
The initial learning rate is $1 \times 10 ^{-3}$ and the cosine schedule was adopted
inspired by \cite{zhang2019bag}.

\begin{figure*}[h]
  \centering
  \includegraphics[scale=0.45]{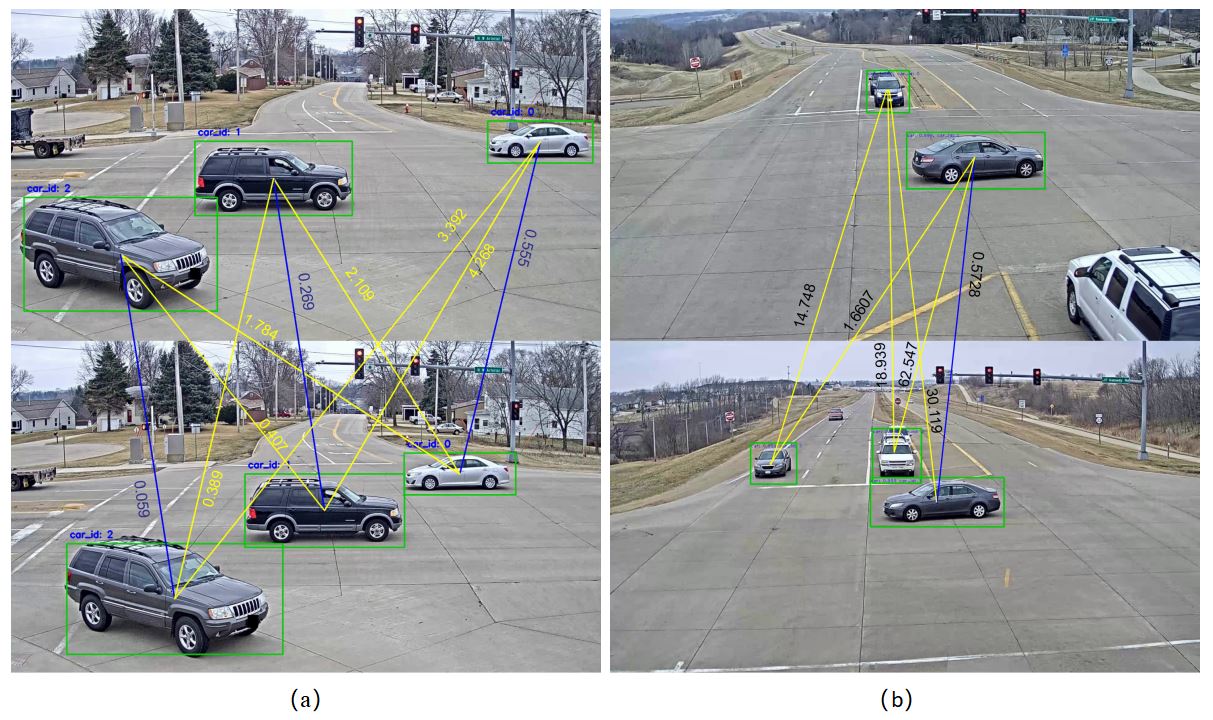}
  \caption{tracking result.
  (a) single camera scenario result, (b) multi-camera scenario result}
  \label{fig:track}
\end{figure*}

\subsection{Inference}
During inference,
We first apply NMS on outputs of detection branch to get final
detect results, we record the preserved index meanwhile. Then select 
corresponding feature by
preserved index and calculate the distance matrix $\mathbb{D}$ between
feature maps of current frame and former frame using trained track branch.
The size of matrix $\mathbb{D}$ is $M \times N$, where $M$ is the 
number of vehicle in current frame, N is the number of vehicle
in former frame.
For $i_{th}$ vehicle in current frame, we get the minimum 
distance $d_{min}^{i}$ and its index $j$ in $i_{th}$ row of matrix $D$.
We first check whether $d_{min}^{i}$ is the
minimum in $j_{th}$ column of matrix $\mathbb{D}$, 
then check whether $d_{min}^{i}$ is lower than distance threshold $h$,
where h is the upper bound of same distance when inference.
Only when $d_{min}^{i}$ satisfied above two condition, the $i_{th}$ vehicle 
of current frame can be thougth as the same with 
$j_{th}$ vehicle in former vehicle.

We minimize the following formula
\begin{equation}
  \centering
  \mathop{\arg\min}_{h} (\frac{fp}{gn} + \frac{fn}{gp})
\end{equation}
on a dev-set to choose the proper distance threshold $h$,
where $fp$, $fn$, $gp$, $gn$ is the number of false positive, 
false negative, ground truth positive and ground truth 
negative respectively under a given $h$.

\subsection{multi object track multi camera}
We also construct a dataset by concatenating two images
that contains same vehicle in different camera.
Training with this dataset can
make our proposed method available to handle multi object track multi-camera (MTMC) scenario.

\section{Experiment}
Our model is trained and evaluated on AIC19\footnote{Dataset
can be downloaded at https://www.aicitychallenge.org/}dataset.
AIC19 dataset contains 36 labeled videos for training 
and doesn't contain validation dataset, so we pick
3 videos for validation and the rest for training.
Our train dataset is constructed on divided
AIC19 train dataset by concatenating the neighbouring frame,
It contains 17556 images and 683 vehicle instances.
We directly use picked 3 videos for validation and
these videos contains 2988 images and 97 vehicle instance.
The MTMC train dataset is constructed on divided AIC19 train
dataset by concatenating images that contain same vehicle
in different cameras and finally contains 10378 images and 169 vehicles.
The validation dataset for MTMC is sample from divided AIC19 validation
dataset, it contains 826 groups and each group contains 2 images that
from different cameras and contain same vehicle. This validation for MTMC
is only to calculate pair accuracy and totally contains 1652 images and 82
vehicles.

\begin{table*}
  \centering
  \begin{tabular}{|c|c|c|c|c|c|c|c|c|c|c|c|c|}
    \hline
    & \multicolumn{6}{c|}{single camera} & \multicolumn{6}{c|}{multi camera }\\
    \hline
    model & TP & TN & FP & FN & GP & GN & TP & TN & FP & FN & GP & GN \\
    \hline
    full & 5176 & 6098 & 2 & 16 & 5335 & 6615 & 645 & 4729 & 1036 & 432 & 1093 & 5953 \\
    \hline
    frozen detect & 4989 & 5700 & 27 & 34 & 5335 & 6615 & 575 & 4350 & 1496 & 495 & 1093 & 5953 \\
    \hline
    frozen same & 5196 & 6088 & 1 & 0 & 5335 & 6115 & 701 & 4667 & 1149 & 366 & 1093 & 5953 \\
    \hline
  \end{tabular}
  \caption{pair accuracy}
  \label{fig:pair_acc}
\end{table*}

\textbf{Detection Result}
We use AP and metrics to evaluate the performance of the detector,
AP represents the average precision rate, which is computed over ten 
different IoU thresholds (i.e.,0.5:0.05:0.95).
Our detector can achieve 57.79 \% mAP.

\textbf{Track Result}
We use Multiple Object Tracking Accuracy (MOTA) to evaluate
the performance of the tracker.
MOTA can be formulated as:

\begin{equation} 
  \centering
  MOTA = 1 - \frac{M + FP + MM}{T}
\end{equation}
where $M$ is sum of misses in all frames,
$M = \sum_{t}m_{t}$.
$FP$ is sum of false positive in all frames, $FP = \sum_{t}fp_{t}$.
$MM$ is sum of mismatch in all frames, $MM = \sum_{t}mm_{t}$
$T$ is sum of the number of objects presents in all frames,
$T = \sum_{t}g_{t}$.

However, there are many unlabeled vehicles in the dataset therefore many detected
vehicles will be thought as false positive, leading to a poor MOTA.
We also use a metric named pair accuracy to evaluate track performance
because of this situation.
To calculate pair accuracy, we first select matched detected vehicle with labeled vehicle
using IoU. Then construct pair connections by track result in 
neighbouring frames and judge whether this connections are
right according to labeled vehicle id.
This metric can reflect track accuracy at every frame.
Pair accuracy is shown in Table \ref{fig:pair_acc}.

The MOTA of our tracker is 8.10 \% and pair accuracy is 99.84 \% when distance thresh $h$ 
is set to 0.343.
The explicit statistic of MOTA is shown in Table \ref{table:mota}. 

\begin{table}
  \centering
  \begin{tabular}{|c|c|c|c|c|}
    \hline
    & FP & missing & miss match & gts \\
    \hline
    full & 604 & 8 & 1 & 667 \\
    \hline
    frozen detect & 585 & 8 & 1 & 667 \\
    \hline
    frozen same & 671 & 7 & 1 & 667 \\
    \hline
  \end{tabular}
  \caption{MOTA}
  \label{table:mota}
\end{table}

\begin{figure*}[htp] 
  \centering
  \includegraphics[scale=0.4]{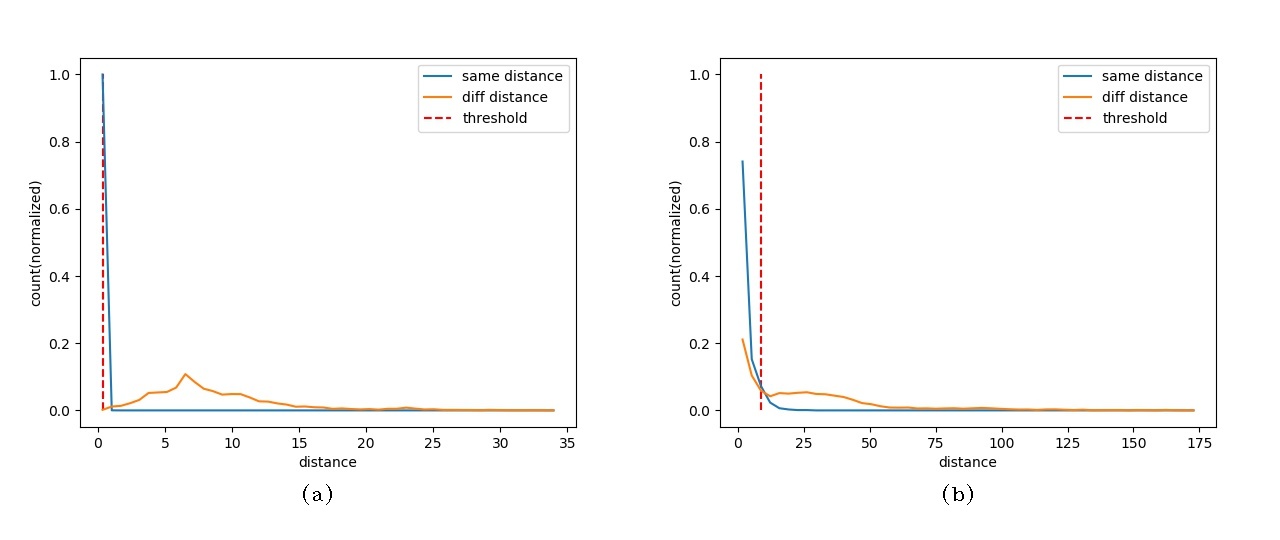}
  \caption{histogram of distance distribution between same objects and different objects.
    (a) is distance distribution in single camera scenario, we choose 0.343 as distance threshold for inference,
  (b) is distance distribution in multi-camera scenario, we choose 8.72 as distance threshold for inference}
  \label{distance}
\end{figure*}

The tracking results are shown in Fig \ref{fig:track}.
The blue lines in the figure represent correct match in 
different frame and yellow lines represent incorrect match.
The number at the top left of detected bounding boxes is the 
predicted tracking id.
The float numbers on the line represent the distances between different
cars in neighbouring frames.
Fig \ref{distance} (a) shows that
distances between same vehicles are much smaller than different vehicles.

\textbf{MTMC}
The images contains same vehicles of different camera can't
be composed to a video, so we only use two images from different camera
to calculate pair accuracy.
The pair accuracy in multi-camera scenario is 78.54 \% when distance thresh $h$ is set to 8.72.

\begin{figure*} 
  \centering
  \includegraphics[scale=0.08]{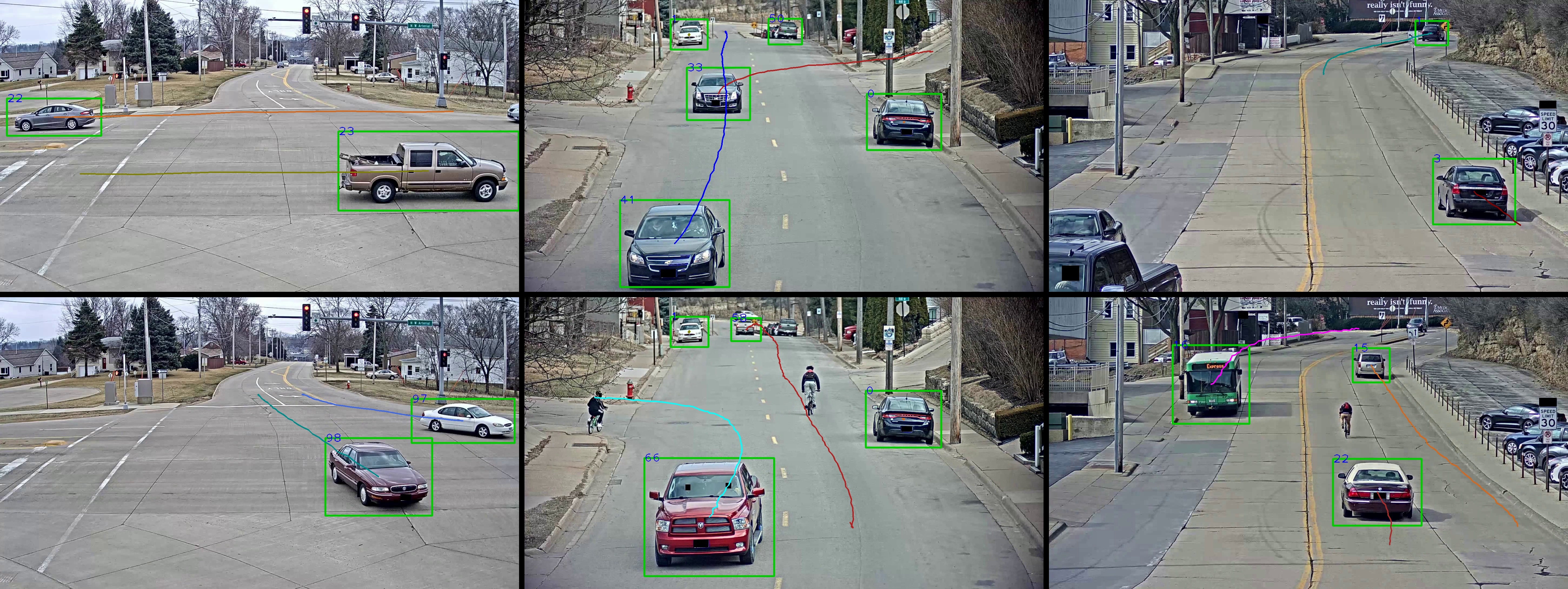}
  \caption{tracking results}
  \label{result}
\end{figure*}

%

\subsection{Ablation Study}

\textbf{multi-task loss} We test the performance between standard Faster 
RCNN and our proposed unified network. The standard Faster RCNN is
trained for 50 epochs on the same dataset.
The standard Faster RCNN can achieve 57.8 \% mAP.

We also test the performance of tracker when we frozen the detector weight.
The MOTA of tracker is 10.94 \% and pair accuracy is 99.43\%
when distance thresh $h$ is set to 3.327.

We test the effect of ``pull" loss by set $\lambda_{4} = 0$.
The MOTA of tracker is -1.80 \% and pair accuracy is 99.99 \%
when distance thresh $h$ is set to 7.529.

\textbf{MTMC}
We evaluate same ablation experiment for multi-camera scenario.
As shown in Table \ref{fig:pair_acc}, pair accuracy of model
that frozen detector in multi-camera scenario is 71.21 \%
when distance thresh $h$ is set to 0.18,
pair accuracy of model that remove ``pull" loss
is 77.99 \% when distance thresh $h$ is set to 28.548.

\subsection{Error Analysis}

There are some match incorrect and track miss situation in our method.
One reason leading to these problems may be that AIC19 dataset didn't label
all the vehicles, but only labeled running car for tracking. It may harm the
training processure and lead to a lower mAP because some still unlabeled vehicles
can be detected when inference.
Another reason may influence track performance is that we only use track feature
to calculate vehicle similarity but not take account of continous coordinate of 
same vehicle.

The reason for lower accuracy in MTMC scenario may be that we don't use center loss
or other metrics to improve the robustness of our model.
Another reason is that the threshold we set to determine whether two vehicle are same
when inference in single camera scenario may not be proper in multi-camera scenario.

\section{Conclusion}
In this work, we have proposed a unified object detection and tracking
network for high quality vehicle detection and tracking.
This method was shown to reduce the amount of computation and make
full use of multi-loss during training and inference.
This is achieved by add an additional sibling track branch to
standard Faster RCNN network, using ROI feature vector as input 
and triplet loss to train this branch.
The dataset needed by training procedure is constructed by
concatenating neighbouring frames in training videos.
Although the not fully labeled dataset had reduced
the metrics of vehicle detection and tracking,
our proposed method can still achieve 57.79 \% mAP and high performance
of vehicle tracking in human eye view.
We believe that our proposed method can be used for many other 
object detection and tracking.

There are many advanced object detection framework in recent few years,
such as Cascade RCNN\cite{cai2018cascade}, FCOS\cite{tian2019fcos}.
We believe that using these better frameworks to replace the Faster RCNN
framework in our work can improve the performance of vehicle detection
and tracking.
we leave as our future work.

\section{Reference}
\bibliography{egpaper_final}
\end{document}